\documentclass[letterpaper, 10 pt, conference]{ieeeconf}  %

\IEEEoverridecommandlockouts                              %

\usepackage{cite}
\usepackage{graphicx} %
\usepackage{siunitx}
\usepackage{subcaption}
\usepackage{xcolor}
\usepackage{tikz}
\usetikzlibrary{shapes.geometric, external}
\usepackage{amsmath} %
\usepackage{amssymb}  %
\usepackage{todonotes}
\usepackage[breaklinks=true]{hyperref}
\usepackage{textcomp}
\usepackage{lipsum}
\newcommand\copyrighttext{%
    \footnotesize \textcopyright © 2024 IEEE.  Personal use of this material is permitted.  Permission from IEEE must be obtained for all other uses, in any current or future media, including reprinting/republishing this material for advertising or promotional purposes, creating new collective works, for resale or redistribution to servers or lists, or reuse of any copyrighted component of this work in other works.
}
\newcommand\copyrightnotice{%
    \tikzset{external/export=false}
    \begin{tikzpicture}[remember picture,overlay]
    \node[anchor=south,yshift=10pt, xshift=10pt] at (current page.south) {\fbox{\parbox{\dimexpr\textwidth-\fboxsep-\fboxrule\relax}{\copyrighttext}}};
    \end{tikzpicture}%
    \tikzset{external/export=true}
}
\definecolor{TUMBlue}{rgb}{0.0, 0.40, 0.74}
\definecolor{TUMGray1}{rgb}{0.2, 0.2, 0.2}
\definecolor{TUMGray3}{rgb}{0.8, 0.8, 0.8}
\definecolor{TUMBlue4}{rgb}{0.60, 0.78, 0.92}
\definecolor{TUMIvory}{rgb}{0.85, 0.84, 0.80}
\definecolor{TUMOrange}{rgb}{0.89, 0.45, 0.13}
\definecolor{TUMGreen}{rgb}{0.64, 0.68, 0.0}
\definecolor{TUMGreenWeb}{rgb}{0.62, 0.73, 0.21}
\definecolor{TUMRedWeb}{rgb}{0.92, 0.45, 0.22}

\title{\LARGE \bf
FlexMap Fusion: Georeferencing and Automated Conflation of HD~Maps with OpenStreetMap}

\author{Maximilian Leitenstern$^{1}$, Florian Sauerbeck$^{1}$, Dominik Kulmer$^{1}$, and Johannes Betz$^{2}$%
\thanks{The research was partially funded by the Federal Ministry of Education and Research of Germany (BMBF) within the project Wies’n Shuttle (FKZ 03ZU1105AA) in the MCube cluster, the Bavarian Research Foundation (BFS), and through basic research funds from the Institute of Automotive Technology.}%
\thanks{$^{1}$Maximilian Leitenstern, Florian Sauerbeck, and Dominik Kulmer are with the Institute of Automotive Technology, Munich Institute of Robotics and Machine Intelligence (MIRMI),
        Technical University of Munich, 85748 Garching, Germany}%
\thanks{$^{2}$Johannes Betz is with the Professorship of Autonomous Vehicle Systems, Munich Institute of Robotics and Machine Intelligence (MIRMI),
        Technical University of Munich, 85748 Garching, Germany} 
\thanks{Corresponding author: \tt\small maxi.leitenstern@tum.de}%
}
\begin{document}
\maketitle
\copyrightnotice
\thispagestyle{empty}
\pagestyle{empty}
\begin{abstract}
Today's software stacks for autonomous vehicles rely on HD maps to enable sufficient localization, accurate path planning, and reliable motion prediction. Recent developments have resulted in pipelines for the automated generation of HD maps to reduce manual efforts for creating and updating these HD maps. We present \textit{FlexMap Fusion}, a methodology to automatically update and enhance existing HD vector maps using OpenStreetMap. Our approach is designed to enable the use of HD maps created from LiDAR and camera data within \textit{Autoware}. The pipeline provides different functionalities: It provides the possibility to georeference both the point cloud map and the vector map using an RTK-corrected GNSS signal. Moreover, missing semantic attributes can be conflated from OpenStreetMap into the vector map. Differences between the HD map and OpenStreetMap are visualized for manual refinement by the user. In general, our findings indicate that our approach leads to reduced human labor during HD map generation, increases the scalability of the mapping pipeline, and improves the completeness and usability of the maps. The methodological choices may have resulted in limitations that arise especially at complex street structures, e.g., traffic islands. 
Therefore, more research is necessary to create efficient preprocessing algorithms and advancements in the dynamic adjustment of matching parameters. In order to build upon our work, our source code is available at \url{https://github.com/TUMFTM/FlexMap_Fusion}.
\end{abstract}
\section{Introduction}
\label{sec:Introduction}%
High-definition (HD) maps are essential to today's autonomous driving systems~\cite{Poggenhans2018, Seif2016}. The term HD map describes a geospatial map for autonomous driving that differs from a conventional, standard-definition (SD) map for human drivers~\cite{Jeong2022}. SD maps can be seen as interactive paper maps with accuracy from \SIrange[]{5}{10}{\meter}, incorporating interactive features and the possibility for map updates. Navigation systems utilize these maps in conjunction with global navigation satellite systems (GNSS) to determine vehicle positions and calculate routes~\cite{Liu2020}. However, due to insufficient accuracy and comprehensiveness for autonomous driving, HD maps have been developed~\cite{Seif2016}. In contrast to SD maps, HD maps provide a three-dimensional representation of the environment with an accuracy of \SIrange[]{10}{20}{\centi \meter}~\cite{Liu2020}. This includes details like lane boundaries, traffic signs, and buildings~\cite{Jeong2022, Bao2022}. Therefore, HD maps are considered an important input for autonomous software stacks, as they are capable of addressing several limitations of visual sensors, such as restricted range and susceptibility to occlusion influences. \\%
\begin{figure}[!t]
	\begin{subfigure}[t]{\linewidth}%
	    \centering%
	    \scalebox{1}{%
	    \includegraphics[width=\linewidth, trim={0 3cm 0 2cm}, clip]{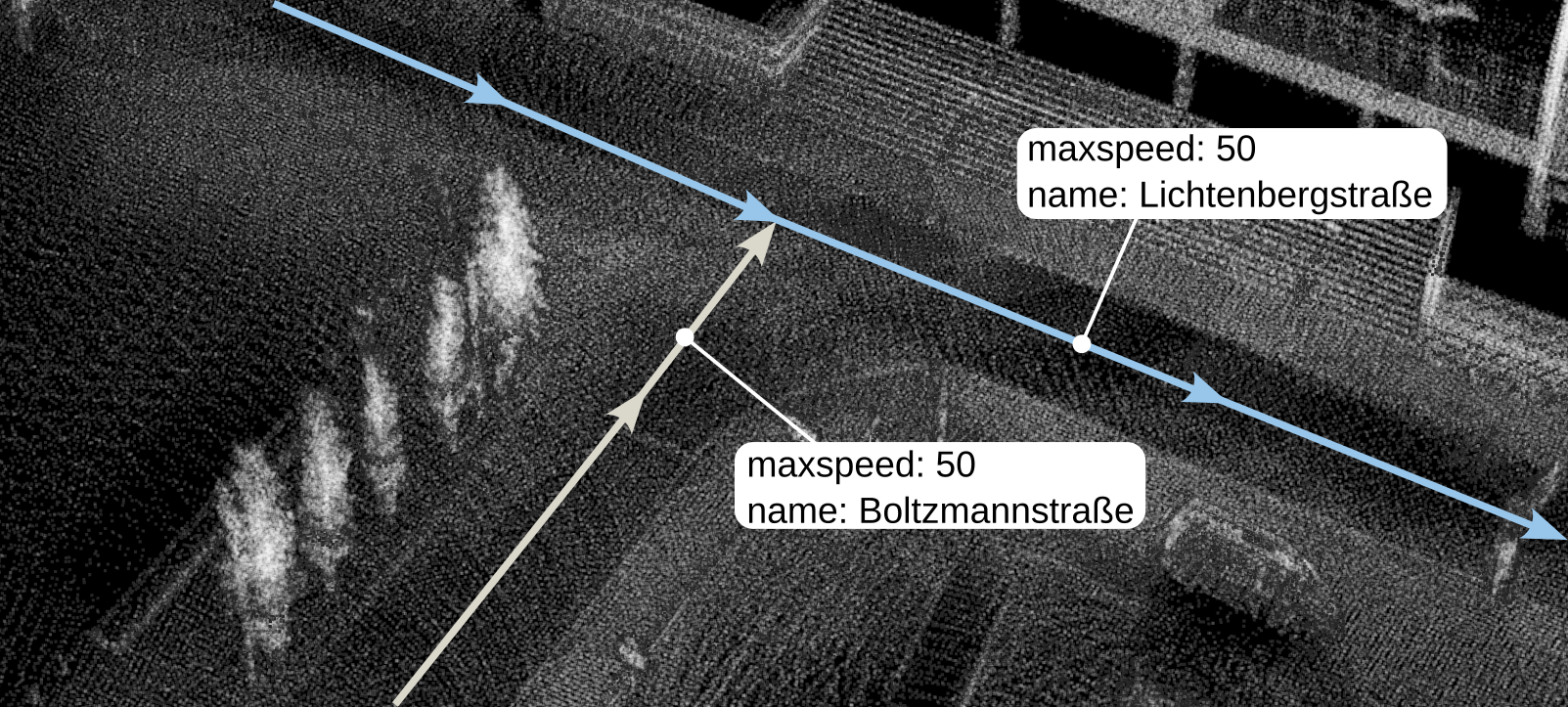}}%
		\label{fig:osm_struc}
	\end{subfigure} \\%
	\begin{subfigure}[t]{\linewidth}%
		\centering%
		\scalebox{1}{%
		\includegraphics[width=\linewidth, trim={0 3cm 0 2cm}, clip]{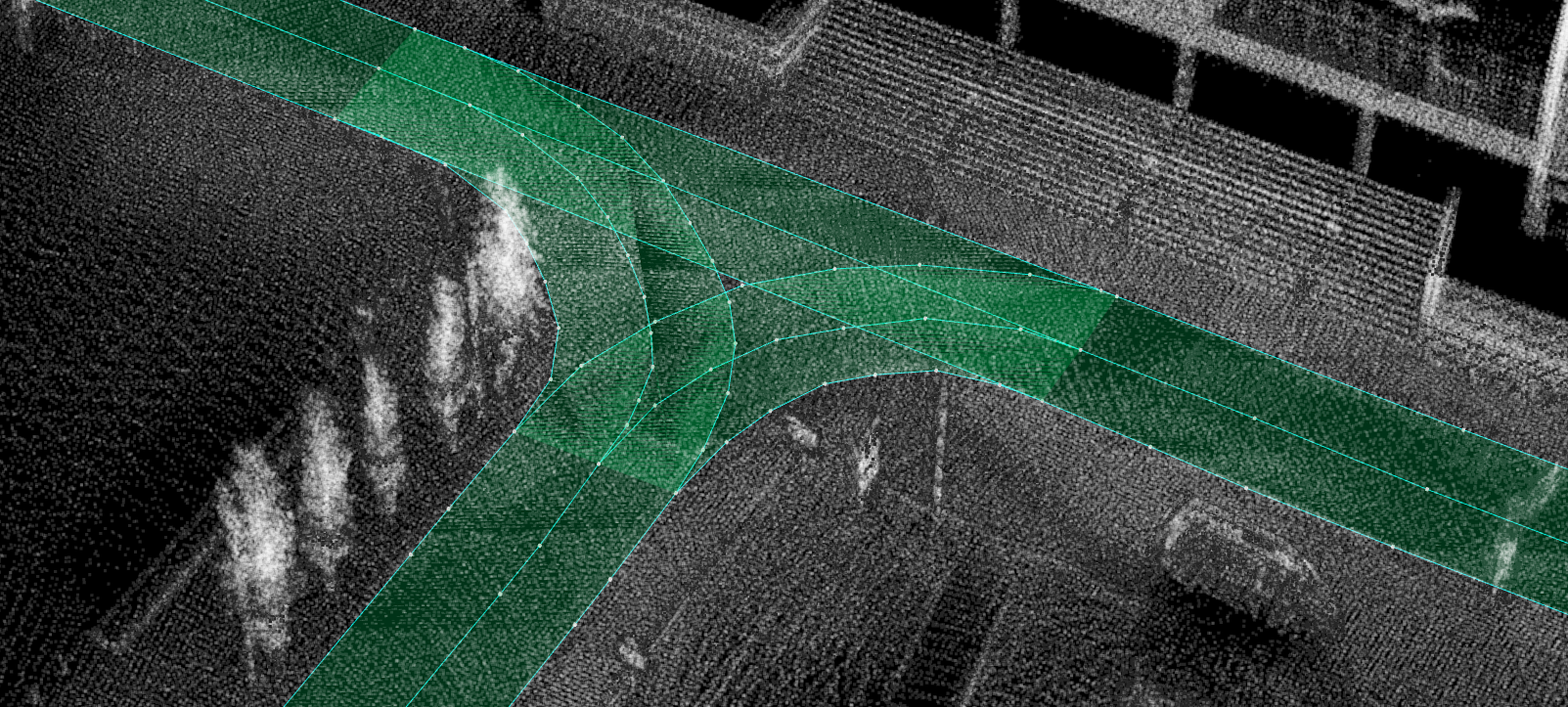}}%
		\label{fig:ll_struc}
	\end{subfigure}
	\caption{PCM with original VM (bottom image) and exemplary information from OSM to be conflated (top image).}%
	\label{fig:road_network}
\end{figure}
Within the field of HD maps, point cloud maps (PCMs) are differentiated from vector maps (VMs). PCMs represent the three-dimensional shape of objects with points, typically generated by registering individual scans from Light Detection and Ranging (LiDAR) sensors~\cite{Jeong2022}. In contrast, VMs describe the georeferenced position of objects of interest, such as lanes, traffic signs, or buildings, by points, lines, and polygons~\cite{Jeong2022}. The traditional, manual annotation of VMs is both costly and time-intensive, leading to scalability issues for large-scale map creation~\cite{Bao2022}. Consequently, the automated generation of large-scale VMs is of high interest. Against this background, the mapping company \textit{HERE}~\cite{HERE} is currently developing \textit{UniMap}, a highly automated mapping technology aiming to create a unified digital map with unique timeliness, quality, and spatial coverage~\cite{James2023}. \textit{UniMap} achieves this by extracting features from sensor data and combining various map formats, such as SD maps or HD maps, into a coherent representation of the environment~\cite{James2023}.

While the feature extraction and fusion of data from various sensors for HD map generation is extensively studied in the literature, few references address the combination of different map formats. To overcome this issue, in this paper, we present with \textit{FlexMap Fusion} a modular approach for postprocessing of existing HD maps that provides the following contributions:
\begin{itemize}
    \item We present a pipeline for automatic georeferencing of PCMs and VMs based on manually selectable control points.
    \item We enable the conflation of semantic attributes from an SD map into a VM.
    \item We provide a validation and visualization of VMs for facilitated manual refinement of the user.
    \item We open-source the pipeline with compatibility to the open-source autonomous driving software stack \textit{Autoware} \cite{Kato2018}
\end{itemize}
\autoref{fig:road_network} visualizes the initial data representation of the VM and \textit{OpenStreetMap} (OSM)\footnote{https://www.openstreetmap.org/} with exemplary attributes that can be conflated using \textit{FlexMap Fusion}.%
\section{Related Work}
\label{sec:RelatedWork}%
In autonomous driving systems, maps address the limitations caused by conventional sensors~\cite{Poggenhans2018}. Real-time access to map information requires a computationally efficient format, leading to the use of HD maps~\cite{Khoche2022}. According to the Automotive Edge Computing Consortium (AECC)~\cite{AECC2020} the required accuracy of a VM is \SIrange[]{10}{20}{\centi \meter}. One of the main challenges in HD map creation, alongside completeness and sufficient accuracy, is to keep the map up-to-date. This requires a modular, scalable framework ensuring easy expandability and modifiability of map information~\cite{Poggenhans2018}. 
The \textit{Autoware} software stack is based on the VM format \textit{Lanelet2}. It was introduced by Poggenhans~et~al.~\cite{Poggenhans2018} in 2018 and uses an XML-based data format with the file extension \textit{.osm} that is also used by OSM. The map structure is decomposed into three layers: The physical layer defines observable objects using points and linestrings. The relational layer connects physical layer elements to lanelets, areas, and traffic rules. Lastly, the topological layer provides contextual information for the relational layer. For data representation, \textit{Lanelet2} contains five so-called primitives:%
\begin{itemize}
    \item \textbf{Points} are the core element defined by 3D position information.
    \item \textbf{Linestrings} are ordered sequences of points.
    \item \textbf{Lanelets} are atomic sections for directed motion, e.g. regular lanes and pedestrian crossings. They are defined by one linestring as left/right boundary. 
    \item \textbf{Areas} are sections for undirected motion, such as parking areas. They are defined by one or more linestrings.
    \item \textbf{Regulatory elements} define traffic rules like speed limits or priority rules. A lanelet/area refers to a regulatory element if it applies to it. 
\end{itemize}
To store the information in the OSM file, the primitives use attributes defining, e.g., the street type or the current speed limit. \par%
Current HD map creation processes involve data recording using a Mobile Mapping System (MMS) with various sensors~\cite{Bao2022}. Subsequently, the sensor data undergoes processing, and geometric features are extracted using vehicle-centric algorithms~\cite{Zheng2019}. Sensor data fusion is studied at various levels to address the limitations of individual sensors and ensure an accurate map. Particularly, approaches focusing on VM creation leverage LiDAR and camera data for map construction~\cite{Li2022, Liu2022, Dong2022, Liao2023}. However, a drawback of these is their limitation to currently extractable information from sensor data. As several features might not be extractable temporarily from sensor data, e.g. due to weather conditions, the incorporation of online map data presents a method for the additional enhancement of VMs. A popular database offering online map data for free is OSM. Several authors successfully leverage OSM for vehicle localization~\cite{Cho2022, Floros2013, Ismail2019}. In the context of map construction, Zhou et al.~\cite{Zhou2021} infer the road network skeleton from OSM to facilitate the construction of a VM. Maierhofer et al.~\cite{Maierhofer2021} present a toolbox for the automatic conversion between OSM, \textit{Lanelet2} and various other map formats. However, our work focuses on combining and validating an existing VM based on OSM. 

The combination of two spatial maps to create an enhanced one is called conflation and has been extensively studied for SD maps like OSM. It usually includes three steps:  During preprocessing, a rubber-sheet transformation removes geometric distortions between the datasets~\cite{Cobb1998}. It is followed by road network matching, which describes the identification of objects in each dataset corresponding to the same object in reality~\cite{Zhang2009}. Lastly, one object of each matched pair is transferred to the improved dataset.

Several authors present methodologies for the conflation of OSM with other SD map formats~\cite{Agarwal2021b, Lei2020, Hacar2019, Zhang2009}. The open-source software Hootenanny\footnote{https://github.com/ngageoint/hootenanny} provides various features for the conflation of different features and maps~\cite{Canavosio2015}. However, these approaches have one thing in common: they do not consider HD map formats. 
\section{Methodology}
\label{sec:Methodology}%
The state-of-the-art developments lack a method to efficiently combine VMs with online map information without much manual effort. We therefore present \textit{FlexMap Fusion} as part of the offline mapping pipeline developed for the EDGAR project at the Technical University of Munich (TUM)~\cite{Sauerbeck2023,EDGAR2024}.

\begin{figure*}
    \centering
    \includegraphics[width=\textwidth]{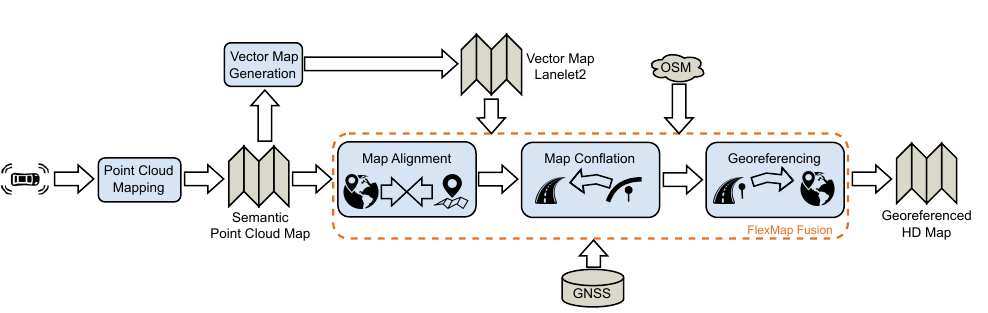}
    \caption{Overview of the presented architecture of \textit{FlexMap Fusion}.}
    \label{fig:architecture2}%
\end{figure*}
As an input, a PCM needs to be created from LiDAR sensor data first. Subsequently, a VM in the \textit{Lanelet2} format is created from the PCM, either manually with \textit{VectorMapBuilder}\footnote{https://tools.tier4.jp/feature/vector\_map\_builder\_ll2/} (chosen method for this work) or automatically using e.g. the approaches by Prochazka et al.~\cite{Prochazka2019}, or Ye et al.~\cite{Ye2022}. As the map creation process does not leverage GNSS, the HD map needs to be georeferenced to enable conflation with OSM data and GNSS usage for localization. Hence, \textit{FlexMap Fusion} consists of three single modules: Firstly, within the \textit{Map Alignment} module, the HD map is aligned to the real-time kinematic (RTK)-corrected GNSS trajectory of the vehicle in a projected, local coordinate frame. Secondly, in the \textit{Map Conflation} module, available data from OSM is conflated into the VM. Thirdly, in the \textit{Georeferencing} module, the projection of local to global coordinates creates a georeferenced VM. The following sections discuss the implementation of the single modules of \textit{FlexMap Fusion} in detail.%
\subsection{Map Alignment} 
The simultaneous localization and mapping (SLAM) process is based only on LiDAR and leads to a vehicle trajectory (from now on referred to as SLAM trajectory) and a corresponding PCM in a local coordinate frame with the initial vehicle position as the origin. The HD map is based on the PCM and thus in the same coordinate system. To enable conflation with OSM data, the maps must be referenced globally and geometric distortions from accumulated errors during the SLAM process must be removed. Belonging to the same vehicle trajectory in reality, the Map Alignment module leverages the SLAM trajectory and a RTK-corrected GNSS trajectory to compute the transformations.\\%
GNSS coordinates and OSM nodes are transformed into a grid coordinate system to facilitate the computation of distances and directions. We use the universal transverse Mercator (UTM) projection provided by the \textit{Lanelet2}\footnote{https://github.com/fzi-forschungszentrum-informatik/Lanelet2} library, which is based on the \textit{GeographicLib}\footnote{https://geographiclib.sourceforge.io/}~\cite{Hager1989}. To limit the shift between the two trajectories, the initial GNSS measurement is selected as the origin for the projection. \\%
Since the orientation of the coordinate system of the SLAM trajectory depends on the initial vehicle heading, the trajectories may still be randomly twisted to each other. Hence, we utilize a rigid transformation to correct for the remaining rotation and potential translation between the trajectories. The algorithm by Umeyama~\cite{Umeyama.1991} computes the least-squares estimation between two sets of points divided into rotation, translation, and scaling. As we do not want to change Euclidean distances within this step, only the rotational and translational parts of the resulting transformation are applied. By applying the transformation to the SLAM trajectory and the HD map, the trajectories and the maps are aligned in the best way without modifying inner structures. \\%
To remove remaining geometric deviations between the trajectories due to accumulated errors during SLAM, we use a piecewise linear rubber-sheet transformation after Griffin and White~\cite{MarvinS.WhiteJr..1985}. Based on a set of so-called control points (corresponding points in each dataset), it applies multiple linear rigid transformations to subdivided regions of the map, specifically triangles. As the transformation matrices of the triangles are derived from the control points defining its corners, only the boundaries are flexed while the transformation stays linear within the triangle~\cite{MarvinS.WhiteJr..1985}. For a detailed explanation on the functionality of a rubber-sheet transformation, the reader is referred to \cite{MarvinS.WhiteJr..1985}. The algorithm is applied to the GNSS and SLAM trajectory of the previous step. At the beginning, the user is prompted to manually select a predetermined number of control points on the two trajectories, showing a high likelihood of correspondence. After the creation of rectangles around the trajectories and their division into two initial triangles, the triangulation is computed. For each control point, the corresponding triangle is determined and replaced by three new triangles. To avoid the creation of long, narrow triangles that may diminish the transformation result, a quadrilateral test is performed on newly created triangles and their neighbors. Subsequently, the single transformation matrices are computed based on the vertices of the defining triangles:
\begin{equation}
    \vec{v_k} = \vec{T_j}\,\vec{u_k}; \quad \vec{v_l} = \vec{T_j}\,\vec{u_l}; \quad \vec{v_m} = \vec{T_j}\,\vec{u_m};
    \label{eq:rs_vertices}
\end{equation}
In this context, $\vec{T_j}$ represents the transformation matrix for a triangle $j$, and $\vec{v_i},\,\vec{u_i}$ with $i \in \{k,\,l,\,m\}$ denote its vertices on the GNSS and SLAM trajectory, respectively. Expanding the equation for an individual vertex yields:
\begin{equation}
    \begin{pmatrix}
        v_{ix} \\
        v_{iy} \\
        1
    \end{pmatrix}
    =
    \begin{pmatrix}
        t_{11} & t_{12} & t_{13} \\
        t_{21} & t_{22} & t_{23} \\
        t_{31} & t_{32} & t_{33}
    \end{pmatrix}
    \,
    \begin{pmatrix}
        u_{ix} \\
        u_{iy} \\
        1
    \end{pmatrix} \quad \quad i \in \{k,\,l,\,m\}.
    \label{eq:rs_matrix}
\end{equation}
Forming this equation for each vertex results in a total of nine equations, enabling the determination of the nine unknown components of $\vec{T_j}$. The transformation is then applied by identifying the surrounding triangle $\vec{T_j}$ of each point $\vec{x}$ and computing
\begin{equation}
    \vec{x}' = \vec{T}_j\,\vec{x}.
    \label{eq:rsPointTransform}
\end{equation}
\autoref{tab:MapAlignmentOverview} summarizes the input parameters of \textit{FlexMap Fusion} with their initial reference system and the applied transformations within the Map Alignment module. As existing approaches for rubber-sheet transformations only work in 2D, the height information of the HD map is currently not adjusted during the Map Alignment process.
\renewcommand{\arraystretch}{1.2}
\begin{table}[!htb]%
    \caption{Overview of Map Alignment process with reference system and applied transforms/projections of inputs.}
    \label{tab:MapAlignmentOverview}
    \centering%
    \begin{tabular}{|p{0.25\linewidth}|p{0.14\linewidth}|p{0.14\linewidth}|p{0.14\linewidth}|p{0.10\linewidth}|}%
        \hline%
        \textbf{Property/ Parameter} & \textbf{GNSS trajectory} & \textbf{SLAM trajectory} & \textbf{PCM/VM} & \textbf{OSM} \\%
        \hline%
        Reference system & global & local & local & global \\%
        \hline
        UTM projection & \checkmark & \text{\sffamily X} & \text{\sffamily X} & \checkmark \\%
        \hline
        Rigid transform & \text{\sffamily X} & \checkmark & \checkmark & \text{\sffamily X} \\%
        \hline
        Rubber-sheet transform & \text{\sffamily X} & \checkmark & \checkmark & \text{\sffamily X} \\%
        \hline%
    \end{tabular}%

\end{table}
\subsection{Map Conflation}
\begin{figure}[!htb]
	\begin{subfigure}[t]{0.5\linewidth}%
	    \centering%
	    \scalebox{1}{%
	    \includegraphics[width=\linewidth]{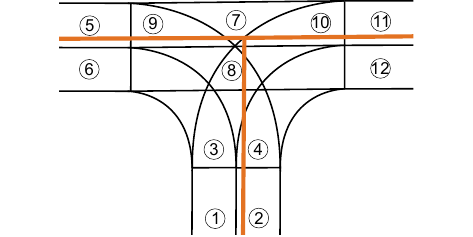}} \\%

	    \footnotesize{~~~~~~~~~~}\rule[0.5pt]{0.5cm}{1pt}\footnotesize{\,: \textit{Lanelet2}~~~~~}\textcolor{TUMOrange}{\rule[0.5pt]{0.5cm}{1pt}}\footnotesize{\,:~OSM}
		\subcaption{intersection representation}
		\label{fig:LoD}
	\end{subfigure}%
	\begin{subfigure}[t]{0.5\linewidth}%
		\centering%
		\scalebox{1}{%
		\includegraphics[width=\linewidth]{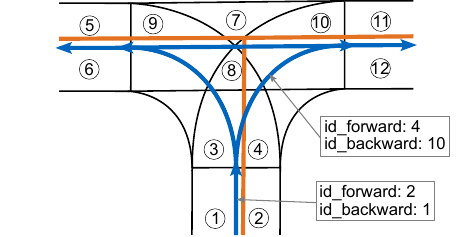}} \\%
		\raggedright
		\footnotesize{~~~~~}\textcolor{TUMBlue}{\rule[0.5pt]{0.5cm}{1pt}}\footnotesize{\,: \textit{Lanelet2} collapsed}
		\subcaption{preprocessed \textit{Lanelet2}}
		\label{fig:collapsed}
	\end{subfigure}
	\caption{Preprocessing of \textit{Lanelet2 map.}}%
	\label{fig:preprocesing}
\end{figure}

\begin{figure}[!htb]
  	\centering
   	\includegraphics[width=\linewidth]{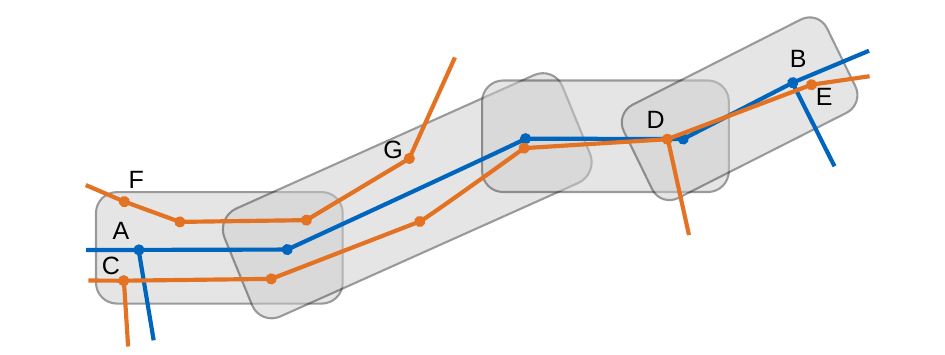} \\%
   	\textcolor{TUMBlue}{\rule[0.5pt]{0.5cm}{1pt}}\footnotesize{\,: \textit{Lanelet2} collapsed~~~~~~~~~~}\textcolor{TUMOrange}{\rule[0.5pt]{0.5cm}{1pt}}\footnotesize{\,: OSM}
   	\caption{Principle of Buffer Growing algorithm.}%
   	\label{fig:bgBuffer}%
\end{figure}

Following the Map Alignment module, the HD and OSM map are geometrically aligned with small deviations. However, due to the SD map format of OSM, it depicts the road network at a different level of detail (LoD) compared to \textit{Lanelet2}~(\autoref{fig:LoD}). Hence, we apply a preprocessing step to the lanelet map as shown in \autoref{fig:collapsed}.
Following the approach of Zhang~\cite{Zhang2009} for matching of dual carriageways, adjacent lanelets are collapsed to a single centerline. A distinction is made between adjacent lanes in the same direction (forward lanes) and adjacent lanes in the opposite direction (backward lanes). The identification numbers of the individual lanelets that a centerline represents are appended to them as attributes. Connecting centerlines based on the connectivity of the lanelets maintains topological consistency to the original map. \\%
For the identification of corresponding objects in the VM and OSM, a road network matching algorithm is implemented. Following the ability to handle different road network resolutions with high matching performance but simultaneously low implementation and parametrization effort, we chose a Buffer Growing algorithm based on Zhang and Meng~\cite{Zhang2005}: First, a reference polyline is created by connecting single centerline segments of the preprocessed lanelet map (e.g. segments $A - B$ in \autoref{fig:bgBuffer}). 
After that, buffers are created around the single segments and all elements from OSM that fall inside them are identified as match candidates (segments $F - G$, $C - D$, $C - E$, and $D - E$ in \autoref{fig:bgBuffer}). If no match candidates could be found, the buffer size is iteratively increased three times. If multiple match candidates are found, the correct match is selected by computing a similarity score between the reference and OSM polyline. The score forms a weighted sum of the angle and distance between a polyline's starting and ending point, its length, and the area of the enclosed polygon. While each segment of the preprocessed lanelet map is only used once during the matching process, OSM segments may be matched multiple times. This addresses sections where the LoD between the lanelet map and OSM differs despite the preprocessing step, e.g. at an intersection (\autoref{fig:preprocesing}). Details on the implementation of the matching algorithm can be found in the work of Zhang and Meng~\cite{Zhang2005}. In contrast to their approach, we do not use self-learning characteristics to improve matching accuracy.\\%
Once corresponding objects are identified in the preprocessed lanelet map and the OSM street network, the available information can be transferred to the lanelet map. As OSM does not contain lane-level details, e.g. the position of road boundaries, direct conflation of positional information from OSM to the lanelet map is not possible. Additionally, the positional accuracy and, notably, the reliability of the OSM road network do not meet the requirements of a lanelet map. Nevertheless, the rich amount of semantic attributes in OSM can be utilized to enhance the lanelet map. Hence, conflation is carried out in three steps:  First, semantic attributes that may not be extractable from LiDAR or camera information are transferred. The attribute \textit{highway} is mapped onto the \textit{Lanelet2} attributes \textit{subtype} and \textit{location} using a custom mapping based on the OSM documentation~\cite{OSM_Wiki}. \autoref{fig:attributeTransferChart} visualizes all implemented correspondences for the Attribute Transfer. 
\tikzstyle{arrow} = [thick,->,>=stealth]
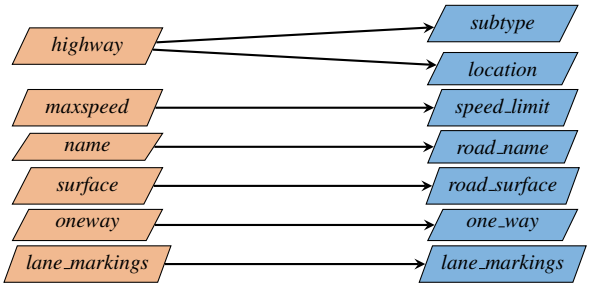
\begin{figure}[!htb]
    \centering
    \begin{tikzpicture}[node distance=0.52cm]
    \node (highway) [trapezium, trapezium stretches=true, trapezium left angle=70, trapezium right angle=110, text centered, minimum width=2cm, minimum height=0.1cm, draw=black, fill=TUMOrange!50] {\footnotesize{\textit{highway}}};
    \node (subtype) [trapezium, trapezium stretches=true, trapezium left angle=70, trapezium right angle=110, text centered, minimum width=2cm, minimum height=0.1cm, draw=black, fill=TUMBlue!50, right of=highway, yshift=0.3cm, xshift=5cm] {\footnotesize{\textit{subtype}}};
    \node (location) [trapezium, trapezium stretches=true, trapezium left angle=70, trapezium right angle=110, text centered, minimum width=2cm, minimum height=0.1cm, draw=black, fill=TUMBlue!50, right of=highway, yshift=-0.3cm, xshift=5cm] {\footnotesize{\textit{location}}};
    \node (maxspeed) [trapezium, trapezium stretches=true, trapezium left angle=70, trapezium right angle=110, text centered, minimum width=2cm, minimum height=0.1cm, draw=black, fill=TUMOrange!50, below of=highway, yshift=-0.3cm] {\footnotesize{\textit{maxspeed}}};
    \node (speed_limit) [trapezium, trapezium stretches=true, trapezium left angle=70, trapezium right angle=110, text centered, minimum width=2cm, minimum height=0.1cm, draw=black, fill=TUMBlue!50, right of=maxspeed, xshift=5cm] {\footnotesize{\textit{speed\_limit}}};
    \node (name) [trapezium, trapezium stretches=true, trapezium left angle=70, trapezium right angle=110, text centered, minimum width=2cm, minimum height=0.1cm, draw=black, fill=TUMOrange!50, below of=maxspeed] {\footnotesize{\textit{name}}};
    \node (road_name) [trapezium, trapezium stretches=true, trapezium left angle=70, trapezium right angle=110, text centered, minimum width=2cm, minimum height=0.1cm, draw=black, fill=TUMBlue!50, right of=name, xshift=5cm] {\footnotesize{\textit{road\_name}}};
    \node (surface) [trapezium, trapezium stretches=true, trapezium left angle=70, trapezium right angle=110, text centered, minimum width=2cm, minimum height=0.1cm, draw=black, fill=TUMOrange!50, below of=name] {\footnotesize{\textit{surface}}};
    \node (road_surface) [trapezium, trapezium stretches=true, trapezium left angle=70, trapezium right angle=110, text centered, minimum width=2cm, minimum height=0.1cm, draw=black, fill=TUMBlue!50, right of=surface, xshift=5cm] {\footnotesize{\textit{road\_surface}}};
    \node (oneway) [trapezium, trapezium stretches=true, trapezium left angle=70, trapezium right angle=110, text centered, minimum width=2cm, minimum height=0.1cm, draw=black, fill=TUMOrange!50, below of=surface] {\footnotesize{\textit{oneway}}};
    \node (one_way) [trapezium, trapezium stretches=true, trapezium left angle=70, trapezium right angle=110, text centered, minimum width=2cm, minimum height=0.1cm, draw=black, fill=TUMBlue!50, right of=oneway, xshift=5cm] {\footnotesize{\textit{one\_way}}};
    \node (lane_markings) [trapezium, trapezium stretches=true, trapezium left angle=70, trapezium right angle=110, text centered, minimum width=2cm, minimum height=0.1cm, draw=black, fill=TUMOrange!50, below of=oneway] {\footnotesize{\textit{lane\_markings}}};
    \node (lane_markings_ll) [trapezium, trapezium stretches=true, trapezium left angle=70, trapezium right angle=110, text centered, minimum width=2cm, minimum height=0.1cm, draw=black, fill=TUMBlue!50, right of=lane_markings, xshift=5cm] {\footnotesize{\textit{lane\_markings}}};
    \draw [arrow] (highway) -- (subtype);
    \draw [arrow] (highway) -- (location);
    \draw [arrow] (maxspeed) -- (speed_limit);
    \draw [arrow] (name) -- (road_name);
    \draw [arrow] (surface) -- (road_surface);
    \draw [arrow] (oneway) -- (one_way);
    \draw [arrow] (lane_markings) -- (lane_markings_ll);
    \end{tikzpicture}
    \caption{Attribute transfer correspondences between OSM (orange) and \textit{Lanelet2} (blue).}
    \label{fig:attributeTransferChart}
\end{figure}

\begin{figure}[!htb]
	\centering
	\includegraphics[width=0.8\linewidth]{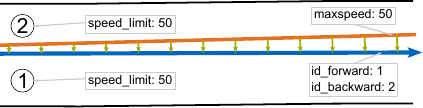} \\%
	\rule[0.5pt]{0.5cm}{1pt}\footnotesize{\,: \textit{Lanelet2}~~}\textcolor{TUMOrange}{\rule[0.5pt]{0.5cm}{1pt}}\footnotesize{\,: OSM~~}\textcolor{TUMBlue}{\rule[0.5pt]{0.5cm}{1pt}}\footnotesize{\,: \textit{Lanelet2} collapsed~~}\textcolor{TUMGreen}{\rule[0.5pt]{0.5cm}{1pt}}\footnotesize{\,: Match links}
	\caption{Principle of attribute transfer.}%
	\label{fig:attributeTransfer}%
\end{figure}

Although there is no counterpart in \textit{Lanelet2}, the OSM tag \textit{lane\_markings} is transferred as is to the lanelet map, as it provides beneficial information for the vehicle when using the map. \autoref{fig:attributeTransfer} visualizes the realization of the attribute transfer based on the matching results. 

Next to the transfer of attributes, the tag \textit{lanes} in OSM is used to validate the correct amount of adjacent lanelets in the lanelet map. Within the visualization of \textit{FlexMap Fusion}, the adjacent lanelets are colored green if the numbers match, red if they differ, and blue if the \textit{lanes} tag is missing in OSM. Lastly, \textit{FlexMap Fusion} can automatically remove mistakenly mapped lanelet fragments from the semantic mapping process (e.g. due to high reflections of objects next to the road). A single lanelet may be deleted if it has neither a predecessor nor a follower (i.e. fragment) and the number of adjacent lanelets at this point exceeds the indicated one from the \textit{lanes} tag in OSM.
\subsection{Georeferencing}
During the Map Alignment process, the SLAM trajectory and the PCM/VM are aligned to the projected GNSS trajectory. Therefore, an inverse application of the UTM projection used in the Map Alignment module enables their georeferencing.
\section{Results}
\label{sec:Results}%
\begin{figure}[!htb]
	\begin{subfigure}[t]{\linewidth}%
	   \raggedright%
	   \scalebox{0.8}{%
	   \hspace{1.5cm}\includegraphics[width=0.82\linewidth]{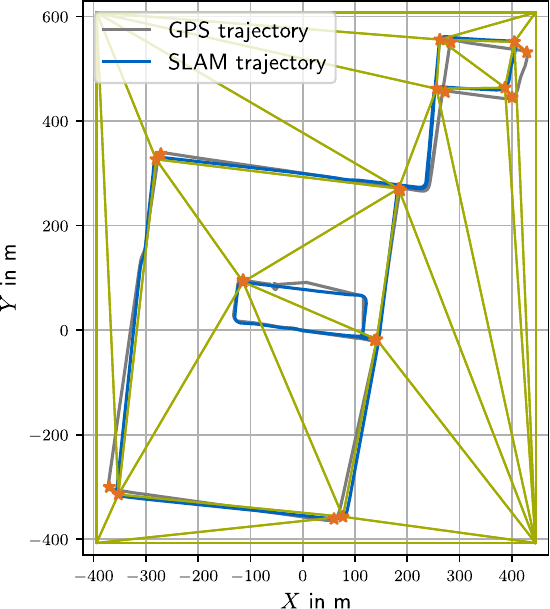}}
	   \subcaption{Triangulation}%
	   \label{fig:geometryRS}%
	\end{subfigure} \\[5pt]%
	\begin{subfigure}[t]{\linewidth}%
		\raggedright%
		\scalebox{0.8}{%
		\hspace{1.5cm}\includegraphics[width=\linewidth]{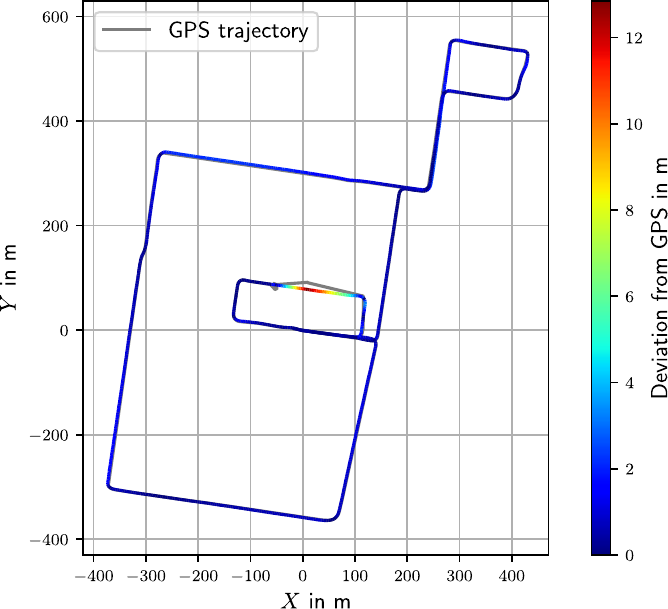}}
		\subcaption{Final alignment}
		\label{fig:trajRS}
	\end{subfigure} 
	\caption{Final results of the Map Alignment process applied to an example route. For triangulation, \SI{10}{} control points are selected manually (orange stars and green triangles). The colored line represents the SLAM trajectory.}%
	\label{fig:trajMatchingRS}
\end{figure}

\begin{figure}[!htb]
	\centering
	\includegraphics[width=\linewidth]{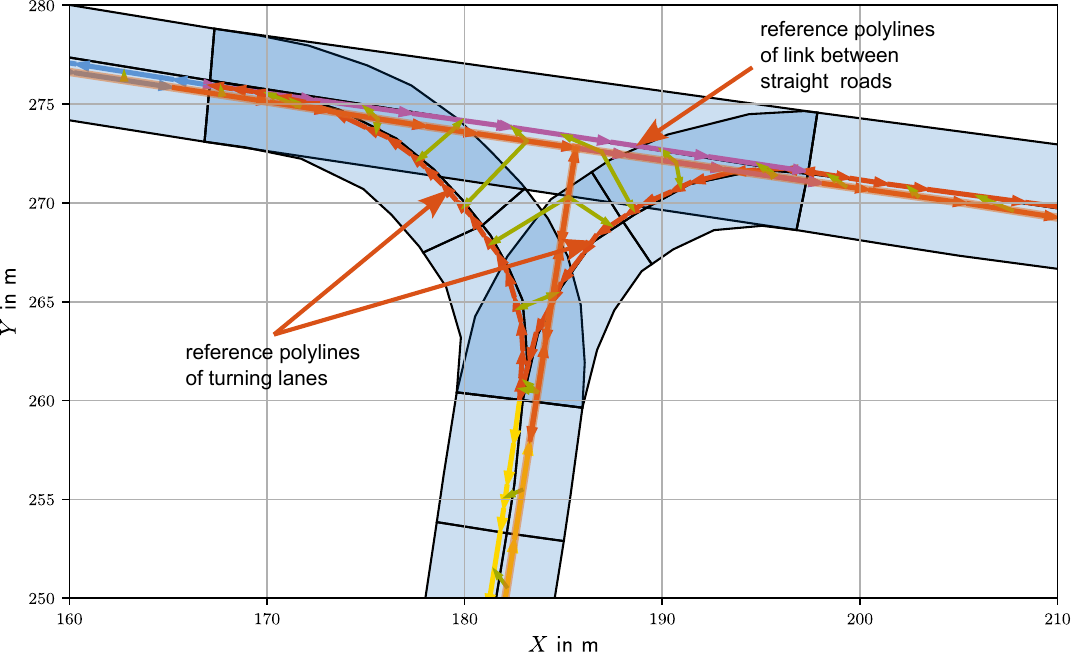} \\%
	\caption{Exemplary matching results for intersection. Green arrows indicate linkages from OSM polylines to reference polylines from the preprocessing step.}%
	\label{fig:matchingIntersection}%
\end{figure}

\begin{figure}[!htb]
	\centering
	\includegraphics[width=\linewidth]{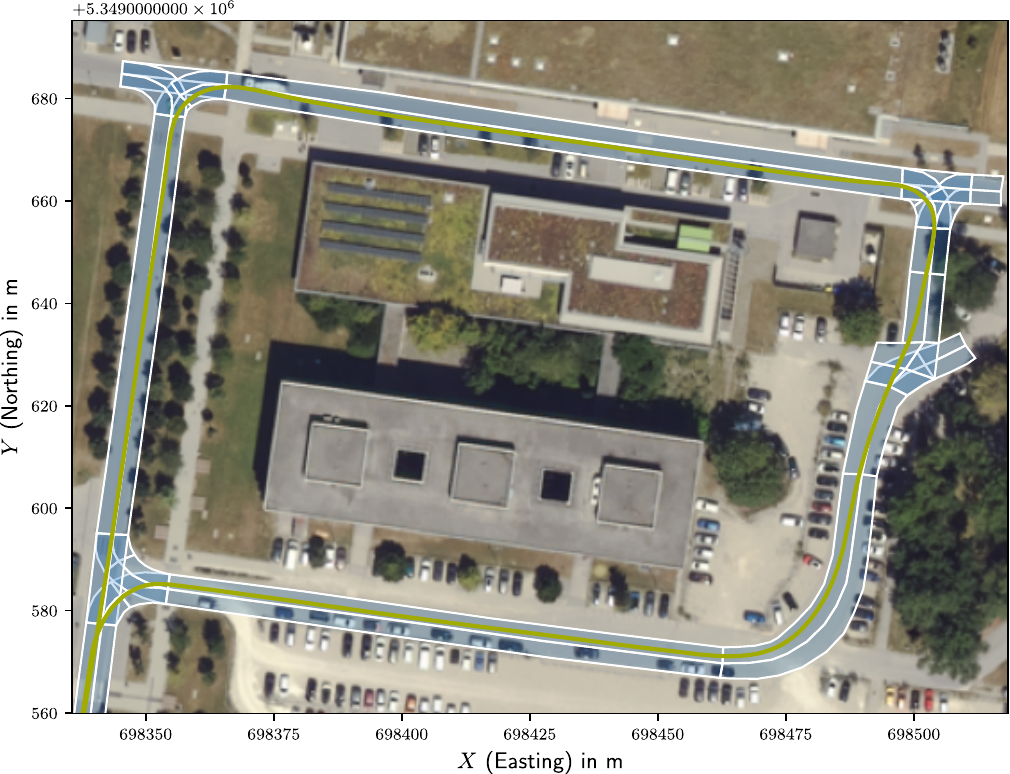} \\%
	\caption{Georeferenced VM and GNSS trajectory over orthophoto of \textit{OPENDATA}.}%
	\label{fig:georef}%
\end{figure}

Experimental results of \textit{FlexMap Fusion} are presented using self-recorded data from the EDGAR research vehicle \cite{EDGAR2024} in Garching, Germany. The initial VMs in this work were created manually as described in \autoref{sec:RelatedWork}. \\%
\autoref{fig:trajMatchingRS} shows the results of the Map Alignment process for a route of EDGAR around the TUM campus in Garching. Despite the Umeyama transformation, deviations from the SLAM trajectory to the projected GNSS trajectory remain (\autoref{fig:geometryRS}). These are removed by the rubber-sheet transformation with manually selected control points (orange stars). As the two trajectories differ in size, the quantitative results are calculated by the minimum distance of a point on the SLAM trajectory to the GNSS trajectory. For the given route and selection of control points, a mean deviation of \SI{1.08}{\meter}, a standard deviation of \SI{1.61}{\meter}, and a root mean squared error (RMSE) of \SI{1.93}{\meter} was achieved. Following the selection of control points, the section with missing GNSS reception (at $X \thickapprox 0$, $Y \thickapprox 100$) does not affect the results negatively \autoref{fig:trajRS}. \\%
The matching algorithm creates a set of reference polylines from the preprocessed lanelet map and searches for matching polylines of OSM segments. This enables a classification into matched and unmatched reference polylines. These categories are further separated by predictive analytics~\cite{Hackeloeer2016}:
\begin{itemize}
    \item \textbf{True Positive} \\%
    Correct identification of match.
    \item \textbf{False Positive} \\%
    Wrong identification of existing match (wrong OSM polyline or no match).
    \item \textbf{True Negative} \\%
    Correct prediction that there is no match.
    \item \textbf{False Negative} \\%
    No identification of an existing match.
\end{itemize}
While the distinction between True and False Positives is based on the geometric similarity score~\cite{Zhang2005}, True and False Negatives need to be identified manually. This allows the computation of precision and recall after~\cite{Hackeloeer2016}. For the given route in \autoref{fig:trajMatchingRS}, our matching algorithm identifies a match for \SI{85.8}{\percent} of the reference polylines with a length above \SI{1.5}{\meter} (the threshold is introduced as short polyline fragments are an unavoidable consequence of the preprocessing step). This leads to a precision of \SI{68.72}{\percent} and a recall of \SI{63.04}{\percent}. \autoref{fig:matchingIntersection} visualizes the matching results for an exemplary intersection.
The algorithm proves to reliably match the created reference polylines to OSM segments despite remaining differences in data representation. Limitations of the matching algorithm are sections with strong differences in road network representation, e.g. at traffic islands that are not present in OSM. Although this may be resolved by increasing the buffer size, it would also increase the amount of False Positive matches. \\%
Finally, the georeferenced lanelet map is to be evaluated. As there is no ground truth data for the position of the road network, the results can only be evaluated qualitatively using orthophotos by \textit{OPENDATA}\footnote{https://geodaten.bayern.de/opengeodata/}. \autoref{fig:georef} shows the results for the top right section of the route from \autoref{fig:trajMatchingRS}.
A visual comparison indicates a strong overlap between the lanelet map and the orthophoto. Within this section, the mean deviation between the SLAM and the GNSS trajectory is \SI{0.38}{\meter} with a standard deviation of \SI{0.27}{\meter} and an RMSE of \SI{0.46}{\meter}. 
\section{Discussion}
\label{sec:Discussion}%
The presented results prove the concept of \textit{FlexMap Fusion}. Our pipeline features a modular structure with strong and reliable algorithms. However, there are still shortcomings in the current implementation that motivate future work. As the control points are selected manually, the results of Map Alignment and thus, Conflation and Georeferencing are not directly reproducible. Furthermore, the user needs reference points for control point selection, e.g. corner apexes. Therefore, deviations between the trajectories increase for sections with few intersections/corners. Also, it needs to be noted that the Map Alignment and Georeferencing processes require a reliable GNSS signal. Taking into account GNSS accuracy and the deviations between the trajectories, our results do not yet fulfill the accuracy requirements of \SIrange[]{10}{20}{\centi \meter} of a VM. However, the accuracy may be increased by automatically selecting control points (e.g. based on corresponding timestamps of GNSS and SLAM poses). \\%
Considering the matching algorithm, problems arise at complex structures where the differences in the LoD of OSM and the preprocessed VM become apparent (e.g. traffic islands that may not be represented in OSM). These problems could be resolved by improving the preprocessing step or a dynamic adjustment of the geometric similarity thresholds (e.g. based on the length of the reference polyline) of the Buffer Growing algorithm. However, it is noted that the evaluation of the matching results may be ambiguous for individual cases. Therefore, additional focus needs to be spent on a clear definition of the four categories defined in \autoref{sec:Results}. Regarding the conflation process itself, larger studies are necessary to evaluate the accuracy and reliability of OSM data. Although the considered area around Garching, Germany showed good coverage in OSM, several sections with missing tags or wrong/ambiguous data were detected. Therefore, a strategy needs to be developed for handling wrong data in OSM. The last point to be mentioned is timeliness: So far, the tool has only been tested with both the VM and OSM being up-to-date. Hence, future work on \textit{FlexMap Fusion} needs to include different temporal versions of map data to account for the case of either one or both maps being outdated. 
\section{Conclusion and Future Work}
With \textit{FlexMap Fusion}, we present a novel approach to enable georeferencing of existing HD maps based on an RTK-corrected GNSS trajectory. Furthermore, our pipeline can automatically enhance and validate VMs using OSM. Although this open-source tool still features limitations, it reduces human labor during HD map generation and thus increases the scalability of the mapping pipeline. Following the exponential growth of users of OSM and thus its reliability and accuracy, work is ongoing on the various approaches to improving the pipeline. 
\bibliographystyle{IEEEtran}
\bibliography{references}
\end{document}